\documentclass[sigconf]{acmart}

\usepackage{booktabs} 

\copyrightyear{2018} 
\acmYear{2018} 
\setcopyright{acmcopyright}
\acmConference[MM '18]{2018 ACM Multimedia Conference}{October 22--26, 2018}{Seoul, Republic of Korea}
\acmBooktitle{2018 ACM Multimedia Conference (MM '18), October 22--26, 2018, Seoul, Republic of Korea}
\acmPrice{15.00}
\acmDOI{10.1145/3240508.3240668}
\acmISBN{978-1-4503-5665-7/18/10}

\usepackage{amsthm}
\usepackage{cases}
\usepackage{bm}
\usepackage{algorithm}
\usepackage{algorithmic}
\usepackage{cases}
\usepackage{subfigure}
\usepackage{multirow}
\usepackage{color} 
\usepackage{enumerate}
\usepackage{etoolbox}
\usepackage{mathrsfs}

\newcommand{\MyMapTemplatePrefix}[4]{\expandafter#1\csname#3#4\endcsname{#2{#4}}}
\newcommand{\MyMapTemplatePrefixNew}[5]{\expandafter#1\csname#4#5\endcsname{#2{#3{#5}}}}
\forcsvlist{\MyMapTemplatePrefix {\def} {\mathbf} {}} {A,B,C,D,E,F,G,H,I,J,K,L,M,N,O,P,Q,R,S,T,U,V,W,X,Y,Z}
\forcsvlist{\MyMapTemplatePrefix {\def} {\mathbf} {}} {a,b,c,d,e,f,g,h,i,j,k,l,m,n,o,p,q,r,s,t,u,v,w,x,y,z,1,0}
\forcsvlist{\MyMapTemplatePrefix {\def} {\widetilde} {wt}} {A,B,C,D,E,F,G,H,I,J,K,L,M,N,O,P,Q,R,S,T,U,V,W,X,Y,Z}
\forcsvlist{\MyMapTemplatePrefix {\def} {\widetilde} {wt}} {a,b,c,d,e,f,g,h,i,j,k,l,m,n,o,p,q,r,s,t,u,v,w,x,y,z} 
\forcsvlist{\MyMapTemplatePrefixNew {\def} {\widetilde}{\mathbf} {tb}} {A,B,C,D,E,F,G,H,I,J,K,L,M,N,O,P,Q,R,S,T,U,V,W,X,Y,Z}
\forcsvlist{\MyMapTemplatePrefixNew {\def} {\widetilde}{\mathbf} {tb}} {a,b,c,d,e,f,g,h,i,j,k,l,m,n,o,p,q,r,s,t,u,v,w,x,y,z}
\forcsvlist{\MyMapTemplatePrefix {\def} {\widehat} {wh}} {A,B,C,D,E,F,G,H,I,J,K,L,M,N,O,P,Q,R,S,T,U,V,W,X,Y,Z}
\forcsvlist{\MyMapTemplatePrefix {\def} {\widehat} {wh}} {a,b,c,d,e,f,g,h,i,j,k,l,m,n,o,p,q,r,s,t,u,v,w,x,y,z}
\forcsvlist{\MyMapTemplatePrefixNew {\def} {\widehat}{\mathbf} {hb}} {A,B,C,D,E,F,G,H,I,J,K,L,M,N,O,P,Q,R,S,T,U,V,W,X,Y,Z}
\forcsvlist{\MyMapTemplatePrefixNew {\def} {\widehat}{\mathbf} {hb}} {a,b,c,d,e,f,g,h,i,j,k,l,m,n,o,p,q,r,s,t,u,v,w,x,y,z}
\forcsvlist{\MyMapTemplatePrefixNew {\def} {\overline}{\mathbf} {lb}} {A,B,C,D,E,F,G,H,I,J,K,L,M,N,O,P,Q,R,S,T,U,V,W,X,Y,Z}
\forcsvlist{\MyMapTemplatePrefixNew {\def} {\overline}{\mathbf} {lb}} {a,b,c,d,e,f,g,h,i,j,k,l,m,n,o,p,q,r,s,t,u,v,w,x,y,z}
\forcsvlist{\MyMapTemplatePrefix {\def} {\mathcal}{mc}} {A,B,C,D,E,F,G,H,I,J,K,L,M,N,O,P,Q,R,S,T,U,V,W,X,Y,Z}
\forcsvlist{\MyMapTemplatePrefix {\def} {\mathbb} {mb}} {A,B,C,D,E,F,G,H,I,J,K,L,M,N,O,P,Q,R,S,T,U,V,W,X,Y,Z}
\forcsvlist{\MyMapTemplatePrefix {\DeclareMathOperator} {} {} } {tr,diag,sgn}
\def\tp{^\intercal} 
\def\ie{{i.e.}} \def\etal{{et.al}}
 \def\eg{{e.g.}}

\allowdisplaybreaks[4]

\begin{document}
\title{Context-Dependent Diffusion Network for Visual Relationship Detection}

\author{Zhen Cui$^\dag$, Chunyan Xu$^\dag$, Wenming Zheng$^\ddag$, and Jian Yang$^\dag$}
\settopmatter{printacmref=false, printfolios=false}

\affiliation{%
	\institution{$^\dag$ PCA Lab, Key Lab of Intelligent Perception and Systems for High-Dimensional Information of Ministry of Education, and Jiangsu Key Lab of Image and Video Understanding for Social Security, School of Computer Science and Engineering, Nanjing University of Science and Technology, Nanjing 210094, China\\
	$^\ddag$ Key Laboratory of Child Development
	and Learning Science of Ministry of
	Education, Research Center for
	Learning Science, Southeast
	University, Nanjing 210096, China}
}
\email{{zhen.cui, cyx, csjyang}@njust.edu.cn; wenming\_zheng@seu.edu.cn}

\renewcommand{\shortauthors}{B. Trovato et al.}

\begin{abstract}
Visual relationship detection can bridge the gap between computer vision and natural language for scene understanding of images. Different from pure object recognition tasks, the relation triplets of subject-predicate-object lie on an extreme diversity space, such as \textit{person-behind-person} and \textit{car-behind-building}, while suffering from the problem of combinatorial explosion. In this paper, we propose a context-dependent diffusion network (CDDN) framework to deal with visual relationship detection. To capture the interactions of different object instances, two types of graphs, word semantic graph and visual scene graph, are constructed to encode global context interdependency. The semantic graph is built through language priors to model semantic correlations across objects, whilst the visual scene graph defines the connections of scene objects so as to utilize the surrounding scene information. For the graph-structured data, we design a diffusion network to adaptively aggregate information from contexts, which can effectively learn latent representations of visual relationships and well cater to visual relationship detection in view of its isomorphic invariance to graphs. Experiments on two widely-used datasets demonstrate that our proposed method is more effective and achieves the state-of-the-art performance. 
\end{abstract}



\maketitle

\textbf{ACM Reference Format:}\\
Zhen Cui, Chunyan Xu, Wenming Zheng, and Jian Yang. 2018. Context-Dependent Diffusion Network for Visual Relationship Detection. In \textit{2018 ACM Multimedia Conference (MM'18), October 22--26, 2018, Seoul, Republic of Korea}. ACM, New York, NY, USA, 8 pages. https://doi.org/10.1145/3240508.3240668

\section{Introduction}

Image understanding is becoming a popular topic in the field of multimedia and computer vision, especially accompanying with recent great successes on the tasks of object detection~\cite{ren2015faster,girshick2015fast} and recognition~\cite{simonyan2014very}. Over past few years, the deep learning techniques have contributed a milestone progress to such basic tasks. In contrast, as a mid-level understanding of visual content, visual relationship detection (VRD) still needs more concerns, since it may effectively underpin high-level image understanding tasks, such as image captioning~\cite{kulkarni2013babytalk,bernardi2016automatic} and question answering~\cite{antol2015vqa,vinyals2017show}.

Visual relationship detection aims to find object pairs of interest and estimate their relations from a given image. We denote a visual relationship with a triplet of \textit{subject-predicate-object}, where the predicate may be spatial, verb, preposition and comparative. Given $N$ objects and $K$ predicates, the combinatorial number of possible relationships is $O(N^2K)$. As a straightforward solution to visual relationship detection, the joint models~\cite{sadeghi2011recognition,ramanathan2015learning} treat each type of triplet as a unique class. However, the long-tailed effect heavily influences the scalability and generalization ability of learned models. To address this problem, the separate models on objects and predicates are often employed in most methods~\cite{lu2016visual,zhang2017visual,li2017vip,dai2017detecting}. In this way, different relation triplets (\eg, \textit{person behind cat}, \textit{house behind car}) are merged
into the same category if they share the same predicate. But the extreme diversity of samples often overwhelms the learning. To tackle this problem, some recent methods attempt to employ language priors~\cite{lu2016visual} or structural learning~\cite{li2017vip, liang2018visual} to reduce unreasonable relationship combinations. However, the relationship triplets are often recognized independently of global contexts, whereas context cues may reduce vagueness of relationships as well as better generalize new relationships.

\begin{figure}[t]
	\centering
	\subfigure[examples of subjects]{\label{fig:sub}
		\includegraphics[width=0.475\linewidth]{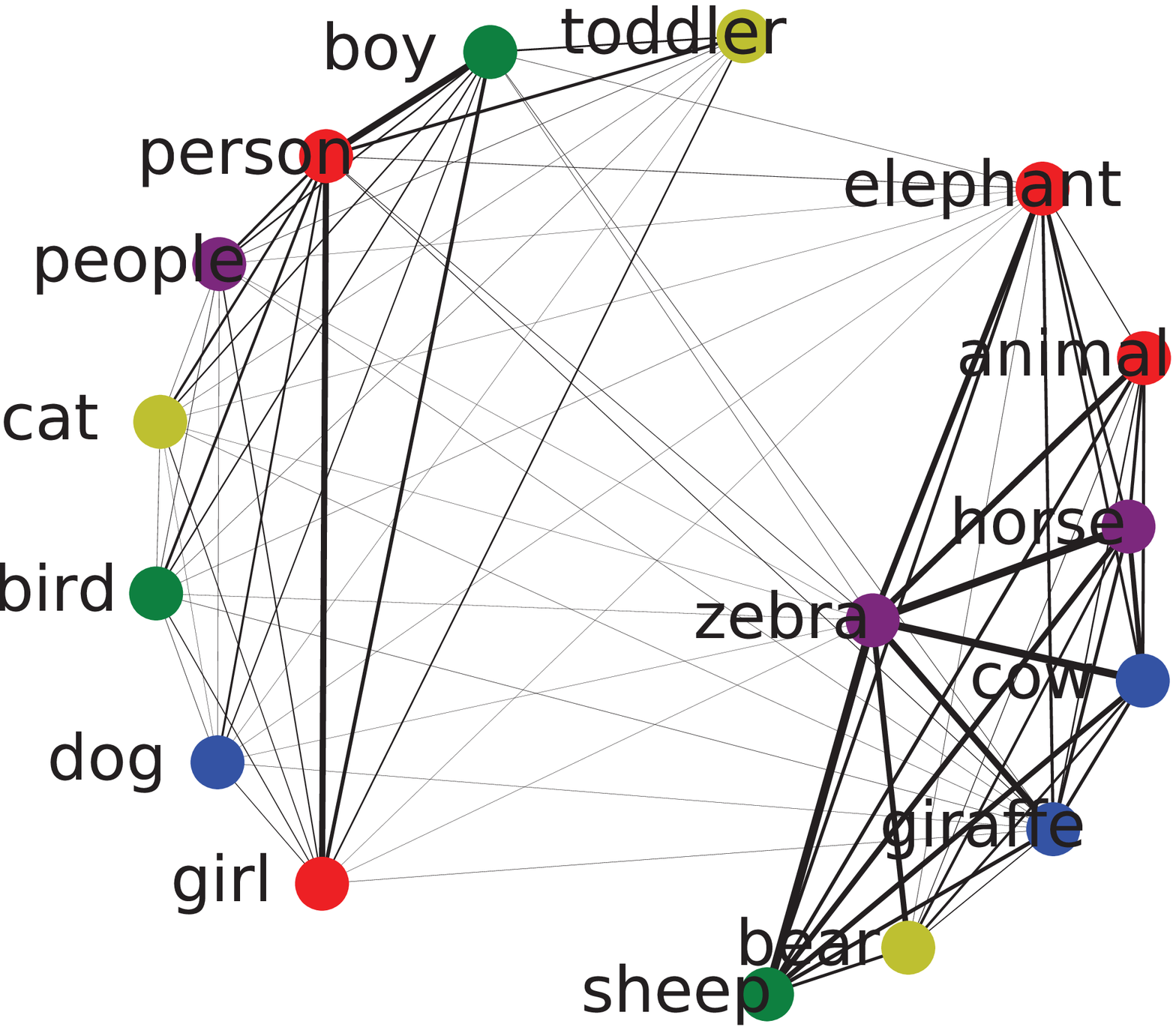}}
	\subfigure[examples of objects]{\label{fig:obj}
		\includegraphics[width=0.475\linewidth]{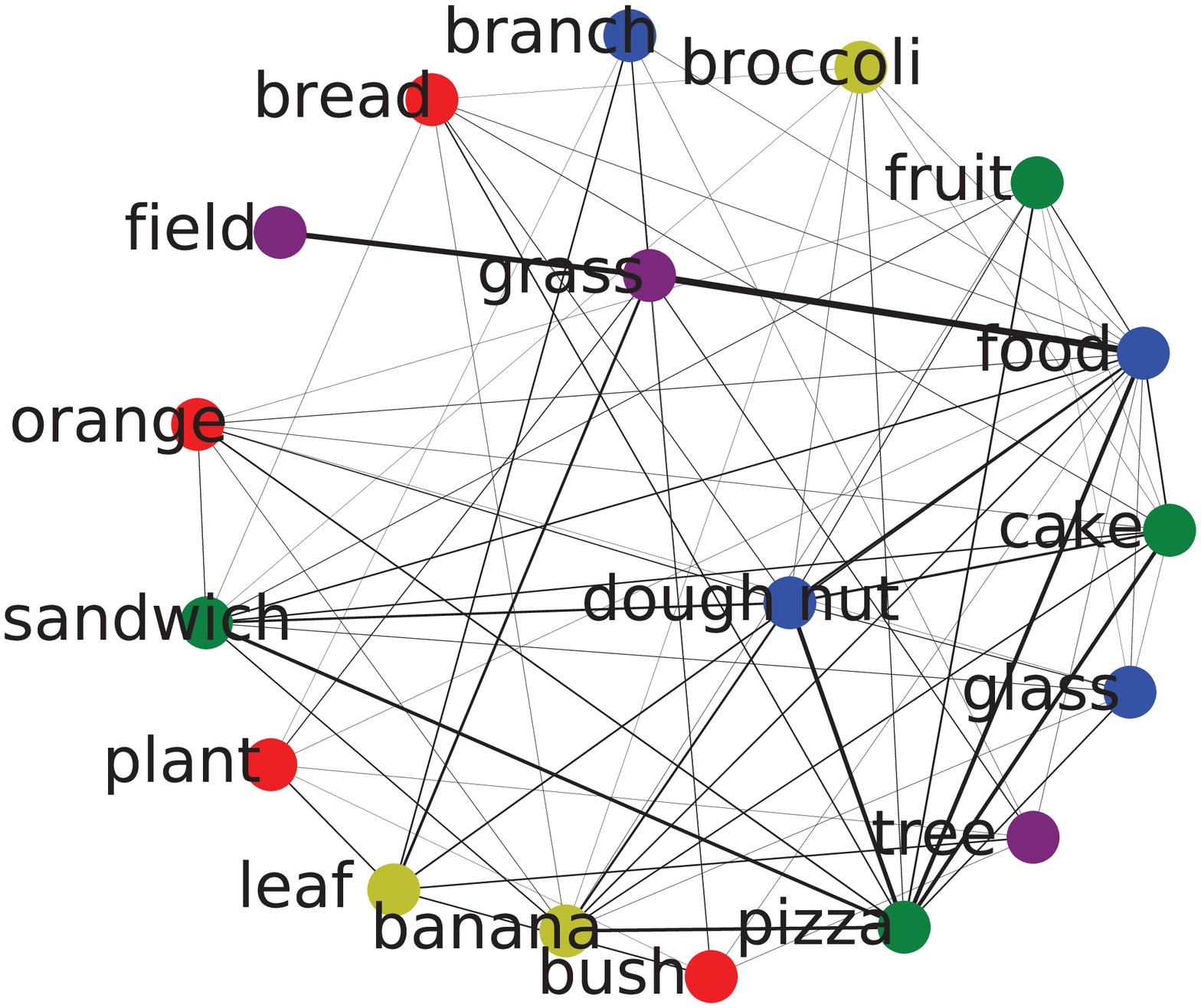} }
	\caption{Motivation. The relationship triplets are often semantically related to each other, \eg, \textit{person-eat-fruit, boy-eat-sandwich, girl-eat-orange, people-eat-fruit}, etc. The produced cliques of \textit{\{person, boy, girl, $\cdots$\}} and \textit{\{sandwich, orange, fruit, $\cdots$\}} should have tight internal correlations, which may be utilized to derive (new) visual relationships. To this end, we build the possibility graph of object semantics by using language priors from training data, and then transmit/exchange information via edge connections, which refers to diffusion in this paper. Note that these two subgraphs only exhibit a few examples for better observation. 
	}
	\label{fig:motivation}
\end{figure}

In this paper, we propose a context-dependent diffusion network (CDDN) framework to deal with visual relationship detection. Our motivation comes from this observation that the subjects/objects are correlated to each other under semantic relationships, as shown in Fig.~\ref{fig:motivation}. Under the shared predicates, some subjects or objects forms some tight cliques, in which connections are with high possibilities. Intuitively, even if some triplets (\eg, \textit{girl-eat-banana}) are not observed never, we can infer them from \textit{person-eat-orange}, according to these two cliques \textit{\{person, boy, girl, $\cdots$\}} and \textit{\{sandwich, orange, fruit, $\cdots$\}}. To this end, we encapsule semantic priors of subjects/objects into semantic graphs by using training data. Given a pair of objects, the information of those related objects is probabilistically transmited to them as an object-related aggregation on semantic graph. Besides, we construct another graph (\ie, visual scence graph) of visual objects based on image scene such that the surrounding context can be captured to promote the performance. Thus, semantic graph is built through language priors to model semantic correlations across objects, whilst visual scene graph defines connections of scene objects by utilizing  surrounding scene information. After creating structured graphs, we design a graph diffusion network to learn latent representations of objects by adaptively gathering context information. The diffusion strategy may be well-catered to visual relationship detection in view of high flexibility of isomorphic invariance. In experiments, we explore different ways to model each input cue and conduct experiments to validate their effectiveness. Experimental results indicate that our proposed method can achieve the competitive performance compared with the state-of-the-art methods on two widely-used datasets: Visual Relationship Dataset (VRD)~\cite{lu2016visual}
and Visual Genome (VG)~\cite{krishna2017visual}.

\section{Related Work}

There are many previous works to tackle the task of visual relationship prediction. At the early stage, only simple relationships are considered, such as spatial predicates (``above", ``inside", ``below" and ``around")~\cite{galleguillos2008object} and human-object interaction relationships~\cite{gould2008multi,yao2010modeling}. The recent deep learning based methods consider more complex situations with more relationships and objects. Generally, these methods fall into two categories: joint model~\cite{ramanathan2015learning,atzmon2016learning} and separate model~\cite{peyre2017weakly,zhuang2017towards,yu2017visual,plummer2017phrase}. The joint models usually take each relationship triplet as a unique class. For example, in the literature~\cite{sadeghi2011recognition}, each type of triplet (\eg,\textit{person-ride-horse}) is trained with a detector. However, the long-tail distribution of visual relation triplets has an inherent defect for model scalability and generalization. Moreover, the combinational number of relationships is explosive if the numbers of objects and relationships are overlarge, so the joint models prefer small-scale relationship datasets.

In contrast, separate models can efficiently alleviate the aforementioned problems. In separate models, subjects, objects and predicates are individually learnt, not jointly learnt on the entire triplets. Thus the complexity can be reduced from $O(N^2K)$ to $O(N+K)$. For example, Lu \etal~\cite{lu2016visual} learned visual models for objects and predicates individually and then combined them together to estimate predicates. To improve the performance of visual modules, they further leveraged language priors through a pre-trained word embedding model. But their network is not an end-to-end manner. To this end, Zhang~\etal~\cite{zhang2017visual} proposed a low-dimensional embedding method of visual relationships by considering the predicate as a translation vector between subject and object. To further establish the connection among relationship components, recently,  Li~\etal~\cite{li2017vip} proposed a visual phrase guided convolutional neural network. Based on spatial configurations and statistical dependencies, Dai~\etal~\cite{dai2017detecting} proposed a deep relational network to associate predicates, subjects and objects. To efficiently combine more cues, Liang~\cite{liang2018visual} proposed a deep structural ranking method for visual relationship detection. In contrast, our method attempts to capture global context cues of between-object interactions. Also different from the recent variation-structured reinforcement learning method~\cite{liang2017deep}, which employed sequential prediction model on semantic action graphs, our method takes a diffusion mechanism on object attribute graphs.

\begin{figure*}[!t]
	\centering
	\includegraphics[width=1.0\linewidth]{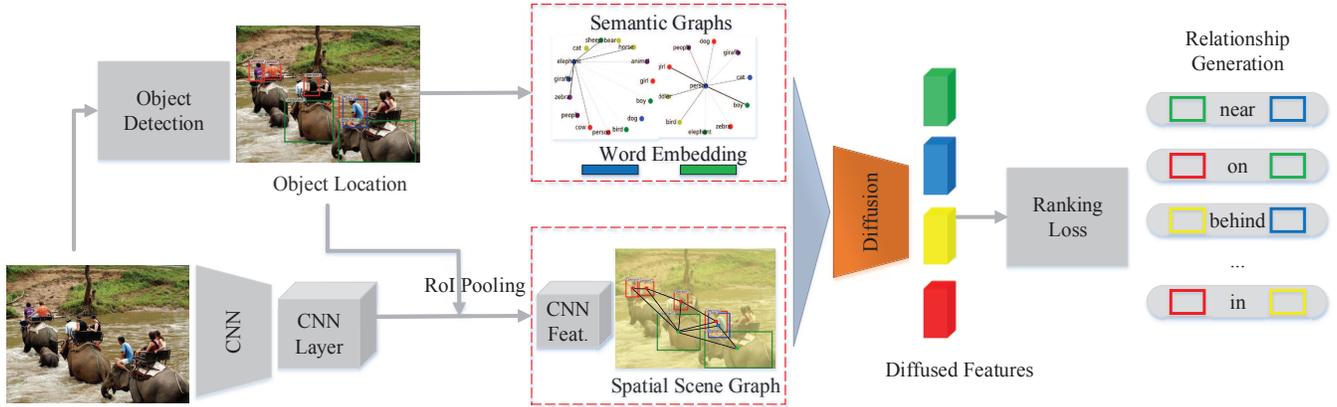}
	\caption{The proposed CDDN architecture of visual relationship detection. We use two types of global context information: semantic priors and spatial scenes. The corresponding graphs are built to capture internal correlations across object instances. The diffusion is used to adaptively propagate context information such that the predicates can be well estimated. Details can be found in Section~\ref{sec:method}.}
	\label{fig:framework}
\end{figure*}

Besides, from the view of methodology, our work is related to representation learning on graphs~\cite{kashima2003marginalized,morris2017glocalized,yanardag2015deep,morris2017glocalized}, especially 
the recent graph convolution methods. Generally, the graph convolution methods fall into two main categories:  spectral methods~\cite{henaff2015deep,kipf2016semi} and spatial methods~\cite{niepert2016learning,li2015gated}. The former often suffer a high-computational burden due to the eigenvalue decomposition of graph Laplacian. A polynomial approximation~\cite{defferrard2016convolutional} may solve this problem to some extent. The latter explicitly model spatial neighborhood relationships through sorting or aggregating neighbor nodes. For examples, diffusion convolution neural network (CNN)~\cite{atwood2016diffusion} performed a diffusion process across each nodes; PSCN~\cite{niepert2016learning} sorted neighbors via edge connections and then performed convolution on sorted nodes; NgramCNN~\cite{luo2017deep} serialized each graph by using the concept of $n$-gram block; GraphSAGE~\cite{hamilton2017inductive} and EP-B~\cite{garcia2017learning} aggregated or propagated local neighbor nodes; WSC~\cite{jiang2018walk} attempted to define ``directional" convolution on random walks by introducing Gaussian mixture models into local walk fields. More recently, Zhao \etal~\cite{zhao2018work} attempted to generalize the ideas/structures of the standard convolution neural network into graph convoution networks.
However, these methods mostly focus on the general graph classification problem. Furthermore, some variants of graph convolution networks are also applied into many fields, such as skeleton-based action recognition~\cite{li2018action,li2018spatio}, good recommendation system~\cite{zhang2018tensor}, EEG-based emotion recognition~\cite{song2018eeg}, and so on. In contrast, in this paper we most focus on the task of visual relationship detection, and intorduce/revise graph convolution (concretely graph diffusion) to cater to such a task.

\section{The Method}\label{sec:method}

In this section, we first overview the entire CDDN architecture, and then introduce four submodules: feature extraction, object association, diffusion layer and ranking loss.

\subsection{Overview}

In visual relationship detection, we need to detect those possible objects and determine the predicates of each pair of them. Let $\mcO$ and $\mcP$ denote the object set and predicate set respectively, then the relationship set can be defined as $\mcR=\{(s,p,o)|s,o\in\mcO, p\in\mcP\}$, where $s$ and $o$ are respectively the subject and the object in a relationship triplet. To better represent objects, we take two types of features: visual appearance and semantic embedding. As shown in Fig.~\ref{fig:framework}, there are two stream pipelines. The top pipeline encodes word semantic features to capture the semantic similarity in language, while the bottom extracts expressive visual appearance features through a convolutional neural network. The diffusion block (Section~\ref{sec:diffusion}) takes two inputs: i) graphs of semantic priors or spatial scenes; and ii) object features of CNN output or word embedding. Given a pair of objects, we can perform the diffusion operation on semantic/spatial scene graphs to integrate global context information. For the construction of graphs, the detailed introduction can be found in Section~\ref{sec:association}. After diffusion, for a pair of objects $(s, o)$, we can obtain the diffused visual features and semantic features for $s, o$. The two types of features are concatenated and then fed into a structural ranking loss to decide the predicates. During the training stage, the relationship triplets are known with pre-annotated objects and relations. In testing, we first perform object detection by some detectors (\eg, faster R-CNN~\cite{ren2015faster}) to acquire the locations, labels and confidence scores for all possible objects, and then send them into this network to estimate the predicates by ranking scores.

\subsection{Features}\label{sec:feature}

Visual appearance plays an important role in distinguishing the categories of objects and understanding relationships. Let $\b_s=(x_s; y_s; w_s; h_s)$ and $\b_o = (x_o; y_o; w_o; h_o)$ denote the tuples of coordinates, width and heights of the detected bounding boxes respectively for a subject and an object. Considering a relative spatial position, we also employ the union bounding box of both regions. Instead of directly using the union, we perform the spatial mask with regard to the subject and object on the union box, as used in the method~\cite{liang2018visual}, such that the relative spatial layouts are preserved. For the subject, the spatial mask suppresses all positions into zeros except for the subject region. Similarly, we conduct the spatial mask for the object region. Thus we can obtain two masked bounding boxes $\b'_s, \b'_o$ on the union of the subject and object. Then we employ convolution neural network as the backbone and use RoI pooling to crop out the features of $\b_s, \b_o, \b'_s, \b'_o$. In our experiments, we choose VGG16~\cite{simonyan2014very}
to extract visual features from the last convolutional layer and then feed RoI features into two fully connected layers.

Language prior is another important strategy to promote visual relationship detection. If only using visual appearance features, the estimation of predicates might be vague sometimes or very difficult due to the large diversity of relationships. Nevertheless, semantic priors can alleviate this problem to some extend and meantime make the inference with a better generality. To use the visual manifestation across different
object categories, we embed each object category into a vector space. Here we use word vector embedding~\cite{mikolov2013efficient} as the off-the-shelf language model to acquire the representations of the subject and object categories. In this way, the links of semantic similarity in language are implicitly encoded. But only this embedding is enough not to characterize relationship datasets, because the embedding model is pre-trained on those general language datasets. It is one reason that we construct semantic graphs of relationships in the next subsection.

\subsection{Object Association}~\label{sec:association}

In order to incorporate global context priors, we construct two types of association graphs on object instances, one is the semantic prior graph and the other is the spatial scene graph. For semantic priors, the visual relationships are semantically related to each other. For example, the triplets, \textit{people riding horse} and \textit{people riding elephant}, have the similar semantics because elephant and horse are both animals. According to human cognition, we should be able to infer the other meaningful triplet, \textit{people riding elephant}, even if we do not observe this case in our life. Thus, under the shared predicates, some object categories are tightly correlated for each other in the semantic space, \eg, \textit{horse, elephant}, even their appearances are rather different. To this end, we expect to build such a graph to make those objects with similar semantics share more information in relationship inference. According to the training data, formally, we define the global semantic graph $\mcG_1=\{\mcV_1,\mcE_1\}$ by scanning all relationship triplets. The node set $\mcV_1$ consists of all candidate object categories, which may be people, places, or animals, etc. The edge set $\mcE_1$ reflects the connection strengthen of object pairs. The larger the edge weight is, the more information the associated objects should share in the inference.  Given two triplets, if their predicates and their subjects (or objects) are consistent, we will set their different objects (or subjects) be connected. For example, the above triplets, \textit{person riding horse} and \textit{person riding elephant}, only have different objects, so the objects \textit{horse} and \textit{elephant} will be connected in the graph. To form a confidence graph, we aggregate all relationships in the training set, and summarize the possibilities of edge connections. Formally,
\begin{align}
	e_{ij} = \frac{1}{\mcN}\sum_{r_1,r_2\in\mcR} \mcI ((r_1 = (s_i,p,o) \wedge r_2 =(s_j, p, o)\wedge s_i\neq s_j) \nonumber\\
	\vee (r_1 = (s,p,o_i) \wedge r_2 =(s, p, o_j)\wedge o_i\neq o_j)),
\end{align}
where $\mcI$ is the indicator function, and $\mcN$ is a normalization factor. 

The other is the spatial scene association on a single image. The main reason why we do scene association is that surrounding scenes usually benefit relationship detection. For scene association, a simple solution is to extract features from an entire image. But it will suffer too background noises irrelevant to the required task. To tackle this problem, we heuristically construct a scene graph $\mcG_2=\{\mcV_2,\mcE_2\}$ for all candidate objects by using their spatial layouts. Given two bounding boxes $\b_i=(x_i; y_i; w_i; h_i)$ and $\b_j = (x_j; y_j; w_j; h_j)$ tied to two objects, we assign their connection score through the following formula,
\begin{align}
	e_{ij} = \left\{\begin{array}{cc}
		1, & \text{if iou$(\b_i,\b_j)>t_1$ or dis$(\b_i,\b_j)< t_2$},\\
		0, & \text{otherwise},
	\end{array}\right.
\end{align}
where $t_1$ and $t_2$ are two threshold values (0.5 as default), ``iou" means the intersection over union of two bounding boxes, and ``dis" is the normalized distance of two boxes. Concretely, we compute the centers of two boxes as $\c_i, \c_j$, and the union of two boxes as $\{l_x,l_y, l_w, l_h\}$. Then the normalized distance is defined as
\begin{align}
	\text{dis}(\b_i,\b_j) = \frac{\|\c_i - \c_j\|_2}{(l_w^2+l_h^2)^{\frac{1}{2}}}.
\end{align}
Therefore, this scene graph $\mcG_2$ mainly defines the visual contextual information while the semantic graph $\mcG_1$ encapsules the semantic correlations across objects. 

\subsection{Diffusion Layer}\label{sec:diffusion}

After object association, we can obtain two types of graphs: semantic prior graph and spatial scene graph. The goal is how to use them to improve the relationship prediction. For the graph-structured data, we introduce the graph diffusion mechanism, which can build latent representation by scanning a diffusion process across each node. In contrast to the graph itself, the graph diffusion can provide a better basis for object representation. More importantly, graph diffusion provides a simple and efficient way to integrate contextual information. Mathematically, the diffusion operation may be formulated as a matrix power series, which may be calculated in polynomial time and efficiently performed on GPU.

Let's retrospect the above constructed semantic graph, where object categories are treated as nodes. If we regard the embedding vector of each object category as the attribute of node, the semantic prior graph is an attribute graph $\widetilde{\mcG}_1=(\mcV, \A, \X)$ of $N$ nodes, where $\mcV=\{v_1,...,v_N\}$ is the set of object categories, $\A$ is the (weighted) adjacency matrix, and $\X$ is the matrix of node attributes. The adjacency matrix $\A\in\mbR^{N\times N}$ records the probability of jumping from one object category to another object category. It can be calculated from the above edge connections $\mcE$. If each object $i$ is endowed with a word embedding vector, \ie, $\x_i:\mcV\rightarrow\mbR^d$, the attributes of all nodes form the attribute matrix $\X=[\x_1,\x_2,\cdots,\x_N]\tp\in\mbR^{N\times d}$, where each row corresponds to one node (\ie, one object catergory). Let $\hbA\in\mbR^{N\times H\times N}$ be a tensor consisting of the power series of $\A$, where the $h$-th slice $\A^h\in\mbR^{N\times N}$ defines the case of the $h$-step hopping starting at each node. Note that the original matrix $\A$ needs to be normalized in advance to avoid the explosion of matrix norm.

Starting at each node $v_i$, the one-hop diffusion process can be formulated as
\begin{align}
	\Z = f(\W\odot\hbA\X, \theta), \label{eqn:fusion}
\end{align}
where $\odot$ represents the element-wise multiplication, $\W\in\mbR^{N\times H\times d}$ is a real-valued weight matrix to be learnt, and $\theta$ is the parameter of the non-linear transformation function $f$. The function $f$ maps a $N\times H \times d$ tensor into another $N\times d'$ tensor after flattening the last two dimensions of the input. Thus, the function $f$ may be a fully connected layer with the network parameter $\theta$. The model in Eqn.~(\ref{eqn:fusion}) comes from the idea of diffusion kernel. As the integration of semantic priors, this model can capture global context priors to some extend. Moreover, the diffusions on two isomorphic input graphs will generate the same activation because of isomorphic invariance of graphs. Finally, we can design a network layer to encapsule the graph diffusion, which can be conveniently incorporated into those general network architectures. 

Similarly, we can perform the diffusion process on spatial scene graphs such that spatial context information can be globally captured to promote the relationship detection.

\subsection{Ranking Loss}

In visual relationship detection, not all predicates are hand-craftly annotated for those object pairs even certain relationship exists between objects. But those annotated relationships should be more salient than those unannotated relationships. Accordingly, we employ the multi-class hinge loss to tackle incompleteness of annotations. Meanwhile, this loss may accommodate relationship to have multiple types of predicates.

Given a pair of objects $s, o$ detected from an input image $x$, we can obtain their representation $\mcF_s,\mcF_o$ respectively after diffusion. Note that each representation $\mcF_s$ or $\mcF_o$ encloses two types of features: diffused visual appearance and diffused semantic embedding. The compatibility function of the relationship triplet $r=(s,p,o)$ can be defined as
\begin{align}
	\psi(r) =  \w_p\tp[\mcF_s;\mcF_o], \label{eqn:comp}
\end{align}
where $\w_p$ is the parameters for the predicate $p$, and $[\cdot,\cdot]$ means the concatenation of two features. Given the relation triplet set $\mcR$, we can construct its complementary set on the predicate space, $\overline{\mcR}=\{(s,p',o)|(s,p,o)\in\mcR, p'\neq p, p'\subseteq\mcP \}$. Then the multi-class hinge loss is formulated as follows,
\begin{align}
	\varsigma(\mcR, \overline{\mcR}) = \sum_{r\in\mcR}\sum_{r'\in\overline{\mcR}}[\epsilon+\psi(r')-\psi(r)]_+,
\end{align}
where $[\cdot]_+=max(0, \cdot)$ takes the positive part of the inputs, and $\epsilon$ is the margin making  positive and negative examples separate asap.

To well adapt to the incompleteness problem of visual relationship detection, we use a dynamic margin by using the prior probabilistic distribution of relationships conditioned on subject and object pairs, as introduced in the literature~\cite{liang2018visual}. Formally, the margin can be defined as
\begin{align}
	\epsilon(r,r') = 1 + P(p|c_s,c_o) - P(p'|c_s',c_o'),
\end{align}
where $c_s, c_o$ denote the class label of subject and object, $P$ is the probability function. Thus, those unannotated pairs with high prior probabilities will have less penalized in the optimization.

During the testing, given any one pair of objects $(s,o)$, we can compute the compatibility score on each predicate $p$ by using Eqn.~(\ref{eqn:comp}). In addition, we also integrate the confidence priors of object detector as well as the distribution prior of the relationship. This strategy can benefit the relation detection as demonstrated in the literature~\cite{liang2018visual}. Formally, let $P(c_s|s), P(c_o|o)$ denote the confidence scores of subject and object produced from the detector, we can define the  conditional probabilistic prior as $P(p|c_s,c_o)P(c_s|s)P(c_o|o)$, where $P(p|c_s,c_o)$ denotes the joint probability prior of subjects and objects on the training set. After aggregating the compatible scores and priors, we can reach the final score of each triplet to be estimated. Finally, we sort all relationship instances and output the top cases as the prediction results.

\section{Implementation Details}

The proposed context-dependent diffusion method employs VGG16 as the base network. We use Adam optimizer to optimize the whole network and the learning rate is set to be 0.00001. In training, the first five convolutional layers of the base network are fixed without finetuning. For those newly added layers,
the learning rate is set to 0.0001 to speed up the training process. We iterate the optimization about 30 epochs by decaying the learning rate with factor 0.1 after each 5 epochs. The implementations use the Pytorch deep learning tools and run on a single GeForce GTX TITAN X. On VRD dataset, the training time will be about several hours without any optimized codes.

\section{Experiments}

We conduct the experiments of our proposed CDDN on two public datasets released recently. First, we give the comparisons with those state-of-the-art methods, then provide an ablation study on different components for our proposed method, finally we show the zero-shot relationship detection to verify the generalization ability on those relationships not occurred in the training data.

\subsection{Datasets and Settings}

Two public datasets, Visual Relationship Dataset (VRD)~\cite{lu2016visual} and Visual Genome (VG)~\cite{krishna2017visual}, are used to test various visual relation detection methods.

Visual Relationship Dataset~\cite{lu2016visual} is a popular dataset for visual relation detection. This dataset consists of 5,000 images with 70 predicates and 100 object categories. It contains 37,993 relation annotations, where the number of unique relations is 6,672 and one object category has about 24.25 predicates on average. Following the previous setting~\cite{lu2016visual}, we split the train and test set with 4,000 images and 1,000 images respectively, where 1,877 relationship triplets only exist in the test set for zero-shot evaluation.

In Visual Genome~\cite{krishna2017visual}, since the annotations of objects and relationships are with much noise in original version, we employ the latest version~\cite{zhang2017visual}, which used official pruning of objects and relations. In summary, this dataset contains 99,658 images with 200 object categories and 100 predicates. There are totally 1,174,692 annotated relation triplets with 19,237 unique ones. In average, each object category has about 57 predicates. Following the same experiment setting to the literature~\cite{liang2018visual}, we split the data into 73,801 images for training and 25,857 images for testing.

In experiments, two relevant tasks are evaluated: predicate detection and relation detection. For relationship detection, the method needs not only predicting the relevant predicates between each pair of objects but also localizing the objects appearing in the image. Following the same evaluation way in the literuature~\cite{lu2016visual}, we use Recall@50 (R@50) and Recall@100 (R@100) as evaluation metrics for relationship detection, where R@K means the fraction of positive predicted relationships in the top K confident predictions for an image. As incompleteness of annotations, \ie, some predictions might not have the ground truth, we don't consider mean average precision (mAP). 

\subsection{Comparison with State-of-the-art}

We compare our proposed method with several methods including JointCNN and JointBox~\cite{zhang2017visual}, VR-V and VR-LP~\cite{lu2016visual}, VTE~\cite{zhang2017visual}, VRL~\cite{liang2017deep}, Vip-CNN~\cite{li2017vip}, DR-Net~\cite{dai2017detecting}, and DSR~\cite{liang2018visual}. JointCNN and JointBox are two baselines. 
JointCNN is a joint model treating each type of triplet as one class. JointBox uses a softmax
classifier to classify joint bounding boxes of subject and object.
VR-V only uses visual appearance model with the R-CNN detector, while VR-LP further combines VR-V language prior by word semantic embedding. VTE is a novel end-to-end visual translation embedding
network designed for visual relation detection. It is fully-convolutional architecture using a softmax
loss function and only rewards the deterministically accurate predicates.
VRL takes variation-structured reinforcement learning to sequentially discover object relationships
in the input image. ViP-CNN introduces Phrase-guided Message Passing Structure (PMPS) to establish
the connection among relationship components so as to jointly consider the relationship learning problems.
DR-Net introduces dual spatial masks for the spatial configurations and meanwhile exploits
the statistical dependencies between objects and relationships. DSR integrates multiple cues for relationship prediction, including appearance, spatial and semantic cues. For our proposed method, we use spatial apperance features and also use word2vec to acquire the initial category embedding.

\begin{table}[h]
	\caption{Performances (\%) on VRD dataset. "-" denotes the results are not reported in the original paper.}
	\label{tab:vrd-cmp}
	\begin{center}
		\begin{small}
			\begin{sc}
				\begin{tabular}{lcccc}
					\toprule	
					\multirow{2}{*}{Methods} &\multicolumn{2}{c}{Predicate Det.} &\multicolumn{2}{c}{Relationship Det.} \\
					&R@50 &R@100 &R@50 &R@100 \\
					\midrule
					JointCNN    &1.47 &2.03 &0.07 &0.09    \\
					JointBox    &25.78 &25.78 &- &-      \\
					VR-V    &7.11 &7.11 &1.58 &1.85     \\
					VR-LP    &47.87 &47.87 &13.86 &14.70    \\
					VTE    &44.76 &44.76 &14.07 &15.20    \\
					VRL    &- &- &18.19 &20.79    \\
					ViP-CNN    &- &- &17.32 &20.01    \\
					DR-Net    &80.78 &81.90 &17.73 &20.88    \\
					DSR       &86.01 &93.18 &19.03 &23.29 \\
					Ours     & \textbf{87.57} &\textbf{93.76} &\textbf{21.46} &\textbf{26.14} \\
					\bottomrule
				\end{tabular}
			\end{sc}
		\end{small}
	\end{center}
\end{table}

The comparison results of predicate and relationship detection are reported in Table~\ref{tab:vrd-cmp}. Our observations are as follows:
\begin{itemize}
	\item  The joint model JointCNN performs the worst. It indicates that jointly training on three components is extremely difficult due to too many triplet categories. In contrast, JointBox achieves a superior performance as the	learning task only aims to classify the predicates of relationship triplets.
	\item The category information and language prior knowledge can benefit the performance. For VR-V
	and VR-LP, we can observe that the gain comes from these priors besides using visual appearance. 
	\item Joint learning on features and visual relationship detection can consistently improve
	the performance. Both VTE and ViP-CNN simultaneously detect objects and
	predict relationships through an end-to-end deep network framework. But with the increasing difficulty of model training, no much gain can be obtained for these multi-task learning methods. The reason might be attributed to the optimization difficulty of model training, which needs to trade off many modules. 
	\item Multi-cue fusion can improve the performance. DSR combines three cues: appearance, spatial mask, and semantic cues. Spatial mask has indicated the improvement due to its matching with the definition of predicates. This is also verified by the two methods, ViP-CNN and DR-Net. Even they all leverage the dependencies between objects and relationships, but DR-Net obtains better performance. 
	\item The performance of object detector is a crucial factor for visual relationship detection, based on the experiment results of predicate and relationship detection. For relationship detection, the object detector must simultaneously satisfy the localization requirement of the subject and the object with a ratio of larger than 0.5 intersection over union (IoU) to ground truth. But it is nontrivial to reach this requirement even for those state-of-the-art object detectors.
	\item Our proposed method is superior to the recent state-of-the-art methods. Especially, for relationship detection, our proposed method further improves the Recall@100 by 3 percent points. With the similar feature with DSR, our proposed method outperforms DSR on all the evaluation metrics. It demonstrates the global context information should not be neglected for visual relationship detection. 
\end{itemize}

\subsubsection*{\textbf{Zero-shot Learning}}~~

As the long tail effect of visual relationships, it is hard to collect all possible relationships. Thus it is important for a model to check its generalization ability on zero-shot learning. To this end, we use 1,877 relationships that only exist in the test set for the VRD dataset. Even if some relationship triplets (\eg, \textit{elephant-stand on-street}) never occur in the training set, we can infer them from the correlated relationships (\eg, \textit{dog-stand on-street}). The comparisons of zero-shot predicate and relationship detection are shown in Table~\ref{tab:zero}. Our proposed method achieves better performances on detecting zero-shot relationships especially on predicate detection. Our method can dramatically improve the performance on predicate detection by about 5 percent points at R@100. In contrast, the relationship detection has a relative slight improvement. The reason should be the influence of detected bounding boxes, which cannot well match the ground truth boxes (IoU > 0.5).

\begin{table}[h]
	\caption{Zero-shot Performances (\%) on VRD dataset. Those methods without reporting the results on zero-shot setting are excluded from comparison.}
	\label{tab:zero}
	\begin{center}
		\begin{small}
			\begin{sc}
				\begin{tabular}{lcccc}
					\toprule	
					\multirow{2}{*}{Methods} &\multicolumn{2}{c}{Predicate Det.} &\multicolumn{2}{c}{Relationship Det.} \\
					&R@50 &R@100 &R@50 &R@100 \\
					\midrule
					VR-V    &3.52 &3.52 &0.67 &0.78   \\
					VR-LP   &8.45 &8.45 &3.13 &3.52   \\
					VTE     &-    &-    &1.71 &2.14   \\
					VRL     &-    &-    &\textbf{7.94} &8.52   \\
					DSR     &60.90 & 79.81 & 5.25 & 9.20 \\
					Ours    &\textbf{67.66} & \textbf{84.00} & {6.40} & \textbf{10.29} \\
					\bottomrule
				\end{tabular}
			\end{sc}
		\end{small}
	\end{center}
\end{table}

\subsubsection*{\textbf{Comparisons on VG}}~~

In addition, we test our method on the VG dataset. Following the same setting to the lieterature~\cite{liang2018visual}, we conduct the experiments on the predicate detection and the corresponding zero-shot testing. The results are shown in Table~\ref{tab:vg}. Note that object detector plays a crucial role in relationship detection due to the matching metric computation of bounding boxes. Thus, we do not compare relationship detection, which need a fair comparison based on the same proposals detected. 

\begin{table}[h]
	\caption{Predicate detection results (\%) on VG dataset.}
	\label{tab:vg}
	\begin{center}
		\begin{small}
			\begin{sc}
				\begin{tabular}{lcccc}
					\toprule	
					\multirow{2}{*}{Methods} &\multicolumn{2}{c}{Predicate Det.} & \multicolumn{2}{c}{Predicate Det(Zero-shot).} \\
					&R@50 &R@100 &R@50 &R@100 \\
					\midrule
					JointBox    &46.59 &46.77 & - & -    \\
					VTE    &62.63 &62.87  & - & - \\
					DSR    & 69.06 & 74.37 & 14.03 & 23.20\\
					Ours   &\textbf{70.42}   & \textbf{74.92} & \textbf{16.19} & \textbf{26.55} \\
					\bottomrule
				\end{tabular}
			\end{sc}
		\end{small}
	\end{center}
\end{table}

\subsection{Ablation Study}

Our proposed network uses the appearance and semantic cues. In this section, we will discuss their effects on the final performance. The comparison results are shown in Table~\ref{tab:abl}. We first conduct an experiment without diffusion, \ie, not considering global context priors. Correspondingly, three tests are performed with appearance cue, semantic cue or both. We also report the predicate detection and relationship detection results on VRD dataset. It can be observed that semantic cue seems more robust than only appearance cue. The reason might be that semantic cue actually covers the categories and confidence scores of objects, which is the higher level features compared to appearance features. By combining both, the performance can be further improved by about 4\% and 1\% for predicate detection and relationship detection. It indicates that the two cues should be complementary for each other to some extend.

\begin{table*}[ht]
	\caption{Ablation Study (\%) on VRD dataset.}
	\label{tab:abl}
	\begin{center}
		\begin{small}
			\begin{sc}
				\begin{tabular}{ccccc|cccc}
					\toprule 
					& & & & & \multicolumn{4}{c}{zero-shot}\\	
					\multirow{2}{*}{Feature} &\multicolumn{2}{c}{Predicate Det.} &\multicolumn{2}{c}{Relationship Det.} &\multicolumn{2}{c}{Predicate Det.} &\multicolumn{2}{c}{Relationship Det.} \\
					&R@50 &R@100 &R@50 &R@100 &R@50 &R@100 &R@50 &R@100 \\
					\midrule
					appearance &  70.07   & 84.61 & 17.91 & 21.56 & 50.89 & 74.08 & 2.75 & 5.33 \\
					semantic    &81.84 &91.46 &18.44 &22.09 & 58.60 & 79.38 & 3.01 & 6.10 \\
					both    &85.28   & 92.87 & 19.96  & 22.62 & 61.42 & 82.03 & 4.09 & 7.01 \\
					\midrule
					diffusion on semantic &86.05 &93.33 &20.80 &25.96 & 65.35 & 83.66 & 5.79 & 9.25 \\
					diffusion on scene &84.86 &92.94 &20.36 &25.00 & 64.32 & 82.63 & 5.44 & 9.60  \\
					diffusion   & 87.57 &93.76 &21.46 &26.14 & 67.66 & 84.00 & 6.40 & 10.29  \\
					\bottomrule
				\end{tabular}
			\end{sc}
		\end{small}
	\end{center}
\end{table*}

Next, we test the influence of context priors. Taking appearance and semantic cues as the baselines, we perform diffusion only on semantic graphs or spatial scene graphs, \ie, ``diffusion on semantic" or ``diffusion on scene" in Table~\ref{tab:abl}. We can observe that integrating context information can further improve the performance. For semantic graphs, given a relationship triplet \textit{person-riding-elephant}, we can borrow the information of relevant objects. For example, the cliques \{\textit{person, people, girl, boy, man,$\cdots$}\} and \{\textit{elephant, horse, zebra, $\cdots$}\} should have high correlations in their internal connections. Thus, for the subject \textit{person} and the object \textit{elephant}  in the triplet \textit{person-riding-elephant}, we can get more shared information based on the graph connections. Intuitively, the shared information may be  features of other objects. More implicitly, some relationships can be transferred into the current triplet. For example, we can infer new relationship triplets, such as \textit{girl-riding-elephant}, \textit{boy-riding-horse}, etc. By considering the current object information, the model can do a proper information aggregation for each object through graph diffusion. It indicates that the global context priors can enhance the generalization ability of visual relationship model to some extent. Further, by integrating two types of global context priors, the performance can be improved to 87.57\% and 21.46\% for the predicate and relationship detection respectively.

\section{Discussion}

The existing visual relationship detection methods~\cite{li2017vip,liang2018visual,liang2017deep,lu2016visual,zhang2017visual}, including our proposed CDDN method, heavily depend on the existing object detection framework. Even though in this paper we more focus on the algorithm of visual relationship discovery, we cannot bypass the effect of object detector. According to the above experiment analysis, even the current state-of-the-art detector cannot yet reach the accuracy of more than 30\%  for the relationship detection as observed in Table~\ref{tab:vrd-cmp} and \ref{tab:zero}. The accuracies of relationship detection are sharply inferior to the corresponding predicate detection, whereas the latter takes ground truth boxes as inputs while the former uses the bounding boxes detected from visual object detectors. Besides, another reason of this serious degradation might be the annotation incompleteness of visual relationship triplets. 

To mitigate this above problem, one possible solution is to borrow text information (\eg, captions) attached to images, and then build image-word pairs. Even though most objects/attributes are not explicitly annotated, one can develop/use some weakly supervised object detection methods to infer those unannotated objects/attributes.
Accordingly, our proposed method may be extended into image-text based relationship detection, where 
word semantic graphs can be constructed by using language prior of text. Thus it should be a potential direction for the future visual relationship detection to which we will put more effort. Excitingly, the recent work~\cite{li2016event} has made a brave stride toward this direction, although it focuses on the techniques of pattern mining from image-caption pairs and totally differs from our technique line. In addition, we believe visual object detectors as well as visual relation datasets (\eg, VG~[8]) will become matured in future.

\section{Conclusion}

In this paper, we proposed a context-dependent diffusion network framework for visual relationship detection. Our purpose was to capture the interactions of different object instances and better perform the relationship inference. To this end, we introduced two types of graphs, word semantic graph and visual scene graph, in order to encode global object interdependency. The semantic graph can model semantic correlations across objects based on language priors, whilst the spatial scene graph can define the connections of scene objects. We further introduced the graph diffusion mechanism to adaptively propagate the information along the edges of graphs. Experimentally, we verified the diffusion effectiveness on global contexts for visual relationship detection. Experiment results on two widely-used datasets also demonstrated that our propose method was more effective and achieved the state-of-the-art performance.

\begin{acks}
	This work was supported by the National Natural Science Foundation of China under Grant Nos. 61602244, 61772276, U1713208 and 61472187, the 973 Program No.2014CB349303, the fundamental research funds for the central universities No. 30918011321 and 30918011320, and Program for Changjiang Scholars.	
\end{acks}

\bibliographystyle{ACM-Reference-Format}
\bibliography{sample-bibliography}

\end{document}